\documentclass[letterpaper]{article} 
\usepackage{aaai2027}  
\usepackage[hyphens]{url}  
\usepackage{graphicx} 
\usepackage{natbib}  
\usepackage{caption} 
\usepackage{amsmath}
\usepackage{algorithm}
\usepackage{algorithmic}

\usepackage{newfloat}
\usepackage{listings}
\DeclareCaptionStyle{ruled}{labelfont=normalfont,labelsep=colon,strut=off} 
\floatstyle{ruled}
\newfloat{listing}{tb}{lst}{}
\floatname{listing}{Listing}

\usepackage{booktabs}
\newif\ifshowappendix
\showappendixtrue

\title{ActSWM: Action-Sensitive World Models for Long-Horizon Planning in Open-World Games}
\author{
Zhenfeng Gan, ZiTong Zeng, Jiajun Cheng, Yeke Song,\\
Yongyi Tang\textsuperscript{\rm *}, Xueqian Wang\textsuperscript{\rm *}
}
\affiliations{\textsuperscript{\rm *}Corresponding authors}

\nocopyright
\begin{document}


\maketitle

\begin{abstract}
Latent world models support efficient model-predictive control by optimizing future control sequences in latent space and replanning in a receding-horizon manner. However, existing latent predictors often lack stable long-horizon rollout ability, and prediction accuracy alone does not ensure that rollouts remain responsive to the actions being planned. We identify \emph{Context Collapse}, a failure mode in which autoregressive latent predictors maintain high similarity to future states while producing nearly indistinguishable futures under different action sequences. To address this issue, we propose ActSWM, an action-sensitive latent world model grounded in a transition-separation principle: a planning-useful latent dynamics model should keep alternative-action futures distinguishable and make the action associated with each local transition recoverable. Under this principle, action sensitivity is enforced as a constraint on latent rollouts rather than treated only as an auxiliary prediction target, encouraging predicted futures to preserve action-dependent differences over long horizons. Across step-drift analysis, closed-loop Minecraft planning, and cross-game local action recovery, ActSWM preserves larger action-dependent rollout gaps than existing baselines, improves task success in long-horizon interactive settings, and enables world-model-based action recovery from offline gameplay videos.
\end{abstract}

\begin{figure*}[t]
    \centering
    \includegraphics[width=0.95\textwidth]{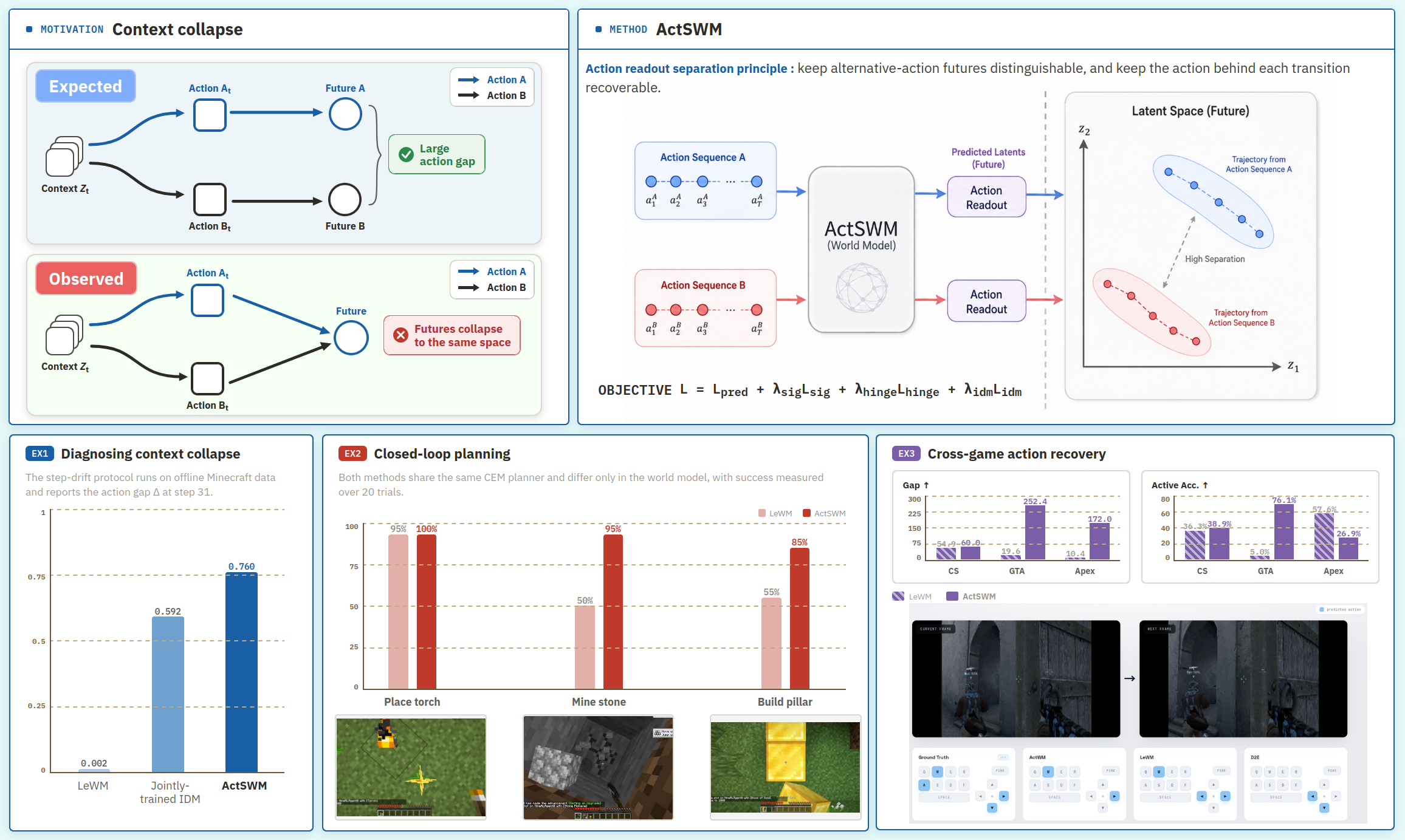}
    \caption{Overview of ActSWM. Latent world models can suffer from context collapse, where long-horizon rollouts match plausible futures but become insensitive to action inputs. ActSWM addresses this by enforcing action-sensitive latent rollouts, leading to larger action gaps, stronger closed-loop planning, and larger CEM-over-random gaps in cross-game local action recovery.}
    \label{fig:overview}
\end{figure*}


\section{Introduction}

Building general-purpose autonomous agents that can perceive, reason, and act in open worlds has long been a central objective in artificial general intelligence (AGI) research. Open-world games provide a concrete testbed for this objective because they combine high-dimensional visual perception, low-level control, and extended task structure. In this setting, recent game agents have made substantial progress: behavior-cloning approaches such as Video Pretraining (VPT) and STEVE-1 learn policies from large-scale gameplay videos \citep{VPT,STEVE1}; multimodal agent systems, including the JARVIS family and Game-TARS, integrate visual perception, language-based reasoning, and low-level control for open-world decision making \citep{JARVIS1,OmniJARVIS,GameTARS}; and Lumine has demonstrated real-time, hours-long task completion in complex 3D open-world environments \citep{Lumine}. Despite these advances, existing methods still face three persistent limitations: low sample efficiency, high deployment cost, and insufficient long-horizon decision-making ability. These limitations become especially pronounced in open-ended tasks that require extended action sequences, such as navigation, resource gathering, object manipulation, and construction \citep{Crafter,MineDojo,Voyager}. In such settings, an agent must not only interpret the current visual state, but also anticipate how actions taken now may cascade into different future states over many time steps.

World-model-based planning offers an attractive route toward addressing these challenges. By learning environment dynamics and optimizing future control sequences in latent space, an agent can evaluate action consequences with substantially fewer real interactions \citep{DreamerV3,TDMPC2}. Recent Joint-Embedding Predictive Architecture (JEPA) methods further suggest that predicting in latent space can improve the efficiency and abstraction of world-model learning \citep{IJEPA,VJEPA}. LeWorldModel (LeWM) applies this idea to lightweight, plannable latent dynamics learned end-to-end from raw pixels \citep{LeWM}. These methods highlight the promise of latent world models for efficient planning and make action sensitivity a central requirement: action conditioning is useful only if distinct action sequences induce separable latent consequences. A latent predictor optimized mainly for future-state similarity does not necessarily satisfy this requirement. The World-Action Model (WAM) augments DreamerV2 with an inverse-dynamics head trained jointly on consecutive encoder embeddings, encouraging the representation to retain action-relevant information \citep{WAM}. However, this objective operates on observed local transitions and does not explicitly constrain predicted rollouts under alternative future action sequences.

We therefore study a rollout-level failure mode that is distinct from the standard accumulation of prediction error in long-horizon world models \citep{ScheduledSampling,HierarchicalLWM,DiffusionForcing}. In this failure mode, degradation arises because action conditioning itself collapses: as the latent context becomes more informative, the predictor can extrapolate state progression from visual history while becoming insensitive to the conditioning action sequence. We refer to this phenomenon as \emph{Context Collapse}. In this setting, longer contexts or multi-step rollout objectives may improve similarity to encoded future states, yet predictions conditioned on recorded actions can remain nearly identical to those conditioned on alternatives such as all-zero actions. For planning-oriented world models, this is particularly harmful: if distinct action sequences fail to induce distinguishable latent futures, a planner cannot reliably compare their downstream consequences.

To address this problem, ActSWM explicitly trains predicted latent rollouts to remain sensitive to the conditioning action sequence. As a result, the predicted futures encode action-dependent state changes rather than merely matching likely future states. This design preserves the efficiency of latent prediction while discouraging action-agnostic rollouts. Figure~\ref{fig:overview} gives an overview of the motivation, method, and evaluation protocol.

Our main contributions are:
\begin{list}{$\bullet$}{%
    \setlength{\leftmargin}{1.2em}%
    \setlength{\labelwidth}{0.7em}%
    \setlength{\labelsep}{0.5em}%
    \setlength{\topsep}{2pt}%
    \setlength{\itemsep}{2pt}%
    \setlength{\parsep}{0pt}%
    \setlength{\partopsep}{0pt}%
}
    \item We introduce \textbf{ActSWM}, a JEPA-based latent world model designed to support both \textbf{long-horizon prediction} and \textbf{action-sensitive rollout dynamics}.
    \item We validate latent world-model planning in open-ended sandbox-game tasks. In MineStudio closed-loop planning, ActSWM improves success over LeWM by up to \textbf{45\%} on stone mining and \textbf{30\%} on pillar building.
    \item To diagnose Context Collapse, we compare long-horizon rollouts conditioned on recorded and zero actions. ActSWM produces a roughly \textbf{380$\times$} larger separation between the resulting futures than the strongest baseline, showing that its rollouts remain responsive to action inputs.
    \item To test whether the learned dynamics can recover controls from offline videos, we use CEM to infer action sequences that reproduce target future frames across three domains. ActSWM achieves up to a \textbf{16.5$\times$} larger cost improvement over random search and improves \textbf{accuracy} by up to \textbf{0.711} in absolute terms.
\end{list}

\section{Related Work}

\subsection{Game Agents}

Game environments are widely used for studying general-purpose agents because they combine visual perception, long-horizon interaction, and reproducible evaluation. Beyond policy pretraining, several lines of work have turned games into broad testbeds for open-ended agent learning. Crafter evaluates diverse survival-game abilities through semantically meaningful achievements, while MineDojo expands Minecraft into a large task suite with internet-scale multimodal knowledge \citep{Crafter,MineDojo}. Another line studies hierarchical decision making: Plan4MC learns reusable Minecraft skills and plans over a skill graph, and Voyager uses an LLM-driven curriculum with an expanding skill library for open-ended exploration \citep{Plan4MC,Voyager}. These works emphasize benchmark construction, external knowledge, high-level skill planning, or language-code agents. Our focus is complementary: we study whether a compact latent world model can preserve action-dependent dynamics so that downstream planners can compare candidate futures reliably.

\subsection{Latent World Models and Planning}

Latent world models support sample-efficient control by learning compact dynamics for planning or policy optimization. PlaNet, Dreamer, and the TD-MPC family combine learned latent transitions with online planning, actor-critic learning, or model predictive control \citep{PlaNet,DreamerV3,TDMPC,TDMPC2}. DIAMOND emphasizes visual fidelity, DINO-WM supports zero-shot goal reaching, and LeWM learns lightweight JEPA dynamics from pixels \citep{DIAMOND,DINOWM,LeWM}. Long-horizon errors arising from exposure bias and accumulated rollout drift have motivated diffusion correction, hierarchical or variable-length prediction, temporal discrimination, and improved transformer training \citep{ScheduledSampling,DiffusionForcing,HierarchicalLWM,VLWM,TDWM,ImprovingTWM}. LS-Imagine further extends imagination for open-world exploration in MineDojo \citep{LSImagine}. These methods improve temporal prediction or policy learning, whereas Context Collapse concerns a distinct axis: accurate rollouts may still respond weakly to alternative actions.

\subsection{Action Representation in Latent Dynamics}

For a world model to support planning, actions must be represented through their effects on latent transitions. Latent-action methods learn controls from passive video or jointly co-evolve latent actions with generative world models, enabling unlabeled data use but introducing an inferred action space \citep{LatentActionWM,CoLAWorld}. WAM, Sensorimotor World Models, and actionable process world models instead use inverse or bidirectional action prediction to preserve locally controllable factors \citep{WAM,SMWM,ActionableWM}. Delta-JEPA decodes actions from latent displacements, while DWM separates action-invariant world effects from action-driven residuals \citep{DeltaJEPA,DWM}. World2Act contrastively aligns policy actions with world-model dynamics for VLA post-training rather than constraining the world model itself \citep{World2Act}. These approaches establish the value of action-aware geometry, while leaving open how to jointly constrain local transitions and multi-step counterfactual futures.

\section{Method}

This section introduces the ActSWM architecture and training objective for learning action-sensitive latent dynamics. ActSWM combines multi-step latent prediction with rollout separation and a fixed action readout that constrains latent transitions to retain recoverable action information.

\subsection{Preliminaries}

Given an offline trajectory of observations and actions $\tau=\{(o_t,a_t)\}_{t=1}^{T}$, our goal is to learn a compact dynamics model that predicts future embeddings rather than reconstructing future pixels. A visual encoder maps observations to $z_t=f_\theta(o_t)$, and an action encoder maps actions to $e_t=g_\theta(a_t)$. The predictor $p_\theta$ takes a window of encoded observations and action features as input and estimates the next latent state:
\begin{equation}
\hat z_{t+1}=p_\theta(z_{t-H+1:t}, e_{t-H+1:t}),
\end{equation}
where $\theta$ collectively denotes the trainable world-model parameters, $H$ is the context length, and $t$ denotes the final time step in the context window. For multi-step prediction, the same predictor is applied recursively under future action sequences, producing latent trajectories for downstream evaluation.

\subsection{Modeling Action Sensitivity}

The formulation above specifies how actions condition the predictor, but conditioning alone does not ensure that distinct action sequences induce distinct rollouts. Action-sensitive latent dynamics should produce distinguishable future predictions whenever different control sequences imply different consequences. We therefore separate action sensitivity into a rollout-level criterion and a transition-level criterion.

At the rollout level, alternative future action sequences starting from the same context should lead to separable predicted futures. For an arbitrary horizon $K$, let $\mathbf{a}^{(1)}_{t:t+K-1}$ and $\mathbf{a}^{(2)}_{t:t+K-1}$ denote two future action sequences, and let $\hat z^{(1)}_{t+1:t+K}$ and $\hat z^{(2)}_{t+1:t+K}$ denote their corresponding $K$-step rollouts from the same context. An action-sensitive world model should keep these futures distinguishable whenever the two sequences imply different consequences. In our training and evaluation, we use the recorded future action sequence and an all-zero sequence as a controlled contrast for measuring this property.

At the transition level, the input action should be recoverable from the change between adjacent latent states. Specifically, a transition $u_t=[z_t,z_{t+1}]$ should contain sufficient information to infer $a_t$, where $[\cdot,\cdot]$ denotes feature concatenation. This criterion is related to inverse-dynamics and bidirectional action-prediction regularization \citep{SMWM,ActionableWM}, while rollout-level separation additionally tests whether action-dependent differences persist over multiple predicted steps.

\begin{figure*}[!t]
\centering
\includegraphics[width=0.9\textwidth]{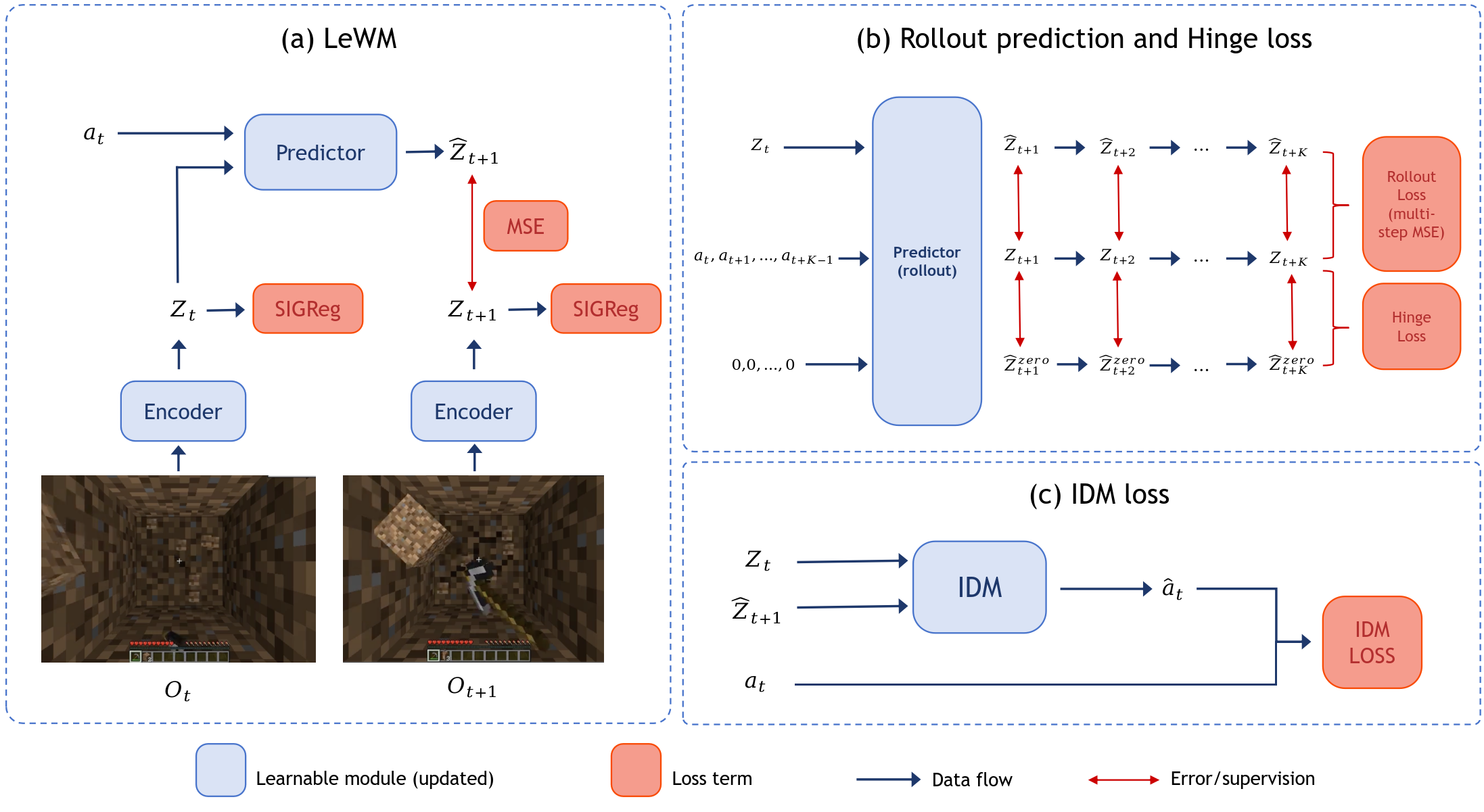}
\caption{ActSWM framework. ActSWM combines JEPA-based latent prediction, multi-step rollout training, an action-sensitivity hinge loss, and a fixed action readout to learn action-sensitive latent dynamics.}
\label{fig:framework}
\end{figure*}

\subsection{Action-Readout Separation Principle}

We implement the transition-level criterion with an action readout. Let $q_{\phi}$ map latent transitions to actions. If the same readout can recover different actions from different latent transitions, then those transitions must be distinguishable in latent space.

This readout-based constraint can be stated in terms of transition separation. Suppose $q_{\phi}$ is locally $L$-Lipschitz and two transitions $u_i,u_j$ are decoded with errors $\epsilon_i$ and $\epsilon_j$, i.e.,
\begin{equation}
\|q_{\phi}(u_i)-a_i\|\le \epsilon_i,
\quad
\|q_{\phi}(u_j)-a_j\|\le \epsilon_j.
\end{equation}
Then the triangle inequality and Lipschitz continuity imply
\begin{equation}
\|u_i-u_j\|
\ge
\frac{\|a_i-a_j\|-\epsilon_i-\epsilon_j}{L}.
\end{equation}
Thus, if different actions are recovered accurately by a shared readout, their latent transitions must remain separated by a positive margin whenever $\|a_i-a_j\|>\epsilon_i+\epsilon_j$. If the readout changes together with the representation, however, the action-recovery loss can decrease by moving the readout boundary rather than by increasing transition separation. ActSWM therefore implements this principle with a parameter-frozen readout, introduced next. Appendix~\ref{app:fixed-readout-separation} gives the complete derivation.

\subsection{ActSWM Architecture}

ActSWM builds on the latent prediction backbone defined above, but augments it with two action-sensitivity mechanisms. The rollout branch compares futures generated by recorded and alternative actions, while the readout branch uses a randomly initialized action readout whose parameters remain frozen during world-model training. The backbone retains a shared prediction space for multi-step prediction and representation regularization. Figure~\ref{fig:framework} illustrates the model architecture.

For an adjacent latent transition $u_t=[z_t,z_{t+1}]$, the action readout $q_{\phi_0}$ predicts the corresponding action $a_t$, where $\phi_0$ denotes its frozen parameters. During training, $\phi_0$ is excluded from optimization, but the readout loss still propagates through its latent inputs. It therefore updates the encoder for encoded transitions and both the encoder and predictor for rollout-predicted transitions. In this way, the fixed readout implements the transition-separation principle above as a constraint on latent transition representations rather than as an adaptive auxiliary decoder.

\subsection{Training Objective}

ActSWM is trained to enforce the rollout-level and transition-level criteria while preserving latent-space prediction fidelity. Given a training segment of length $H+K$ whose context window ends at $t$, the model encodes $z_{t-H+1:t}$ as context and autoregressively predicts the next $K$ latent embeddings using the recorded future actions, producing $\hat z^{\mathrm{gt}}_{t+1:t+K}$. Here, $K$ denotes the training rollout horizon. This gives the multi-step prediction loss
\begin{equation}
\mathcal{L}_{\mathrm{pred}}
= \frac{1}{K}\sum_{k=1}^{K}
\left\|\hat z^{\mathrm{gt}}_{t+k}-z_{t+k}\right\|_2^2.
\end{equation}
This prediction term aligns the rollout conditioned on recorded actions with the encoded future trajectory, but by itself it does not enforce the rollout-level criterion defined above. We therefore add a contrastive rollout from the same initial context, replacing the future actions with zeros to obtain $\hat z^0_{t+1:t+K}$. The hinge loss then penalizes excessive similarity between the futures produced under recorded and all-zero actions:
\begin{subequations}
\begin{alignat}{2}
&\ell_k
&&= \max\!\left(0,
\cos(\hat z^{\mathrm{gt}}_{t+k},\hat z^0_{t+k})-(1-m)\right), \label{eq:hinge-step}\\
&\mathcal{L}_{\mathrm{hinge}}
&&= \frac{1}{K}\sum_{k=1}^{K}\ell_k, \label{eq:hinge-average}
\end{alignat}
\end{subequations}
where $\cos(\cdot,\cdot)$ denotes cosine similarity and $m$ is the hinge-margin hyperparameter. This action-contrastive hinge term encourages the predicted futures to remain separable under distinct action conditions.
Unlike cross-modal policy alignment in World2Act~\citep{World2Act}, this contrast is applied within the world model between futures generated from the same context under different controls.

For the transition-level criterion, we apply the frozen readout to both the encoded transition and the rollout-predicted transition associated with the same action. To express the predicted rollout uniformly from its observed starting point, define
\begin{equation}
\tilde z_{t+k} =
\begin{cases}
z_t, & k=0,\\
\hat z^{\mathrm{gt}}_{t+k}, & 1\le k\le K.
\end{cases}
\end{equation}
For $k=0,\ldots,K-1$, the encoded and predicted transitions are
\begin{equation}
u^{\mathrm{enc}}_{t+k}=[z_{t+k},z_{t+k+1}],
\quad
u^{\mathrm{pred}}_{t+k}=[\tilde z_{t+k},\tilde z_{t+k+1}].
\end{equation}
With the frozen action readout $q_{\phi_0}$, the readout loss is
\begin{subequations}
\begin{alignat}{2}
&\ell^{\mathrm{enc}}_k
&&= \left\|q_{\phi_0}(u^{\mathrm{enc}}_{t+k})-a_{t+k}\right\|_2^2,
\label{eq:readout-encoded}\\
&\ell^{\mathrm{pred}}_k
&&= \left\|q_{\phi_0}(u^{\mathrm{pred}}_{t+k})-a_{t+k}\right\|_2^2,
\label{eq:readout-predicted}\\
&\mathcal{L}_{\mathrm{readout}}
&&= \frac{1}{K}\sum_{k=0}^{K-1}
\left(\ell^{\mathrm{enc}}_k
+\alpha_{\mathrm{pred}}\ell^{\mathrm{pred}}_k\right),
\label{eq:readout-total}
\end{alignat}
\end{subequations}
where $\alpha_{\mathrm{pred}}=1$ in the default configuration.
The first term aligns encoded transitions with the recorded action, and the second applies the same constraint to rollout transitions. Since $q_{\phi_0}$ is fixed, gradients from this loss are propagated only through its latent inputs, thereby updating the encoder and predictor but not the readout parameters. The full training objective is
\begin{equation}
\mathcal{L}
= \mathcal{L}_{\mathrm{pred}}
+ \lambda_{\mathrm{sig}}\mathcal{L}_{\mathrm{sig}}
+ \lambda_{\mathrm{hinge}}\mathcal{L}_{\mathrm{hinge}}
+ \lambda_{\mathrm{readout}}\mathcal{L}_{\mathrm{readout}},
\end{equation}
where $\mathcal{L}_{\mathrm{sig}}$ denotes the SigReg loss adopted from LeWM~\citep{LeWM}, which stabilizes the scale and distribution of prediction-space features, and the $\lambda$ coefficients weight the corresponding objectives. Together, these terms optimize prediction fidelity, alternative-action rollout separation, and action-discriminative transition representations. Training configurations, loss weights, and implementation details are provided in Appendix~\ref{app:training-config}.

\section{Experiments}

\begin{figure*}[!t]
\centering
\includegraphics[width=0.9\textwidth]{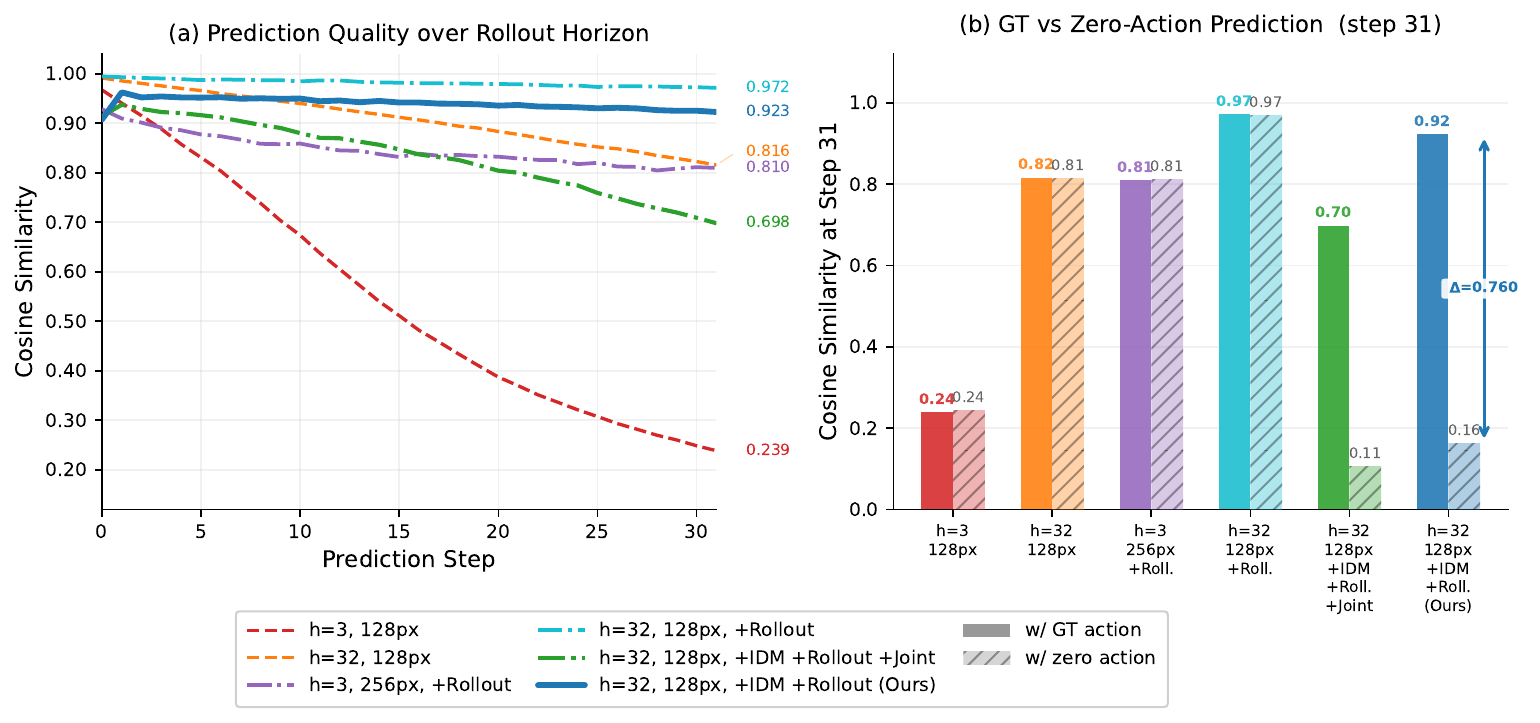}
\caption{Q1 offline step-drift diagnostic on Minecraft VPT trajectories. The left panel tracks recorded-action prediction quality over a 32-step rollout. The right panel compares recorded-action and all-zero-action similarity at step 31 for the context, resolution, rollout-training, and readout variants described in the text. A larger gap indicates stronger action sensitivity.}
\label{fig:step-drift}
\end{figure*}

We organize the experiments around three questions:

\noindent\textbf{Q1: Does ActSWM mitigate Context Collapse?} We test whether predicted rollouts remain distinguishable under different action sequences. Specifically, we compare rollouts conditioned on recorded and all-zero actions to determine whether the model relies on action inputs rather than context alone.

\noindent\textbf{Q2: Does action-sensitive latent prediction improve closed-loop planning?} We evaluate whether a model that better separates action consequences in latent space can guide closed-loop control optimization more reliably in interactive tasks.

\noindent\textbf{Q3: Are the learned action representations more discriminative?} We evaluate local action recovery using the CEM-over-random action gap, overall key accuracy, and active-key accuracy on offline gameplay videos.

\subsection{Q1: Diagnosing Context Collapse}

\noindent\textbf{Setup and variants.} The evaluation uses offline Minecraft H5 trajectories from VPT~\citep{VPT}, where each window provides time-aligned image frames and low-level actions. The main setting uses context length $H=32$, rollout horizon $K_{\mathrm{eval}}=32$, 128px inputs, and a frameskip of 4. We compare LeWM variants with short context ($H=3$), 256px inputs, and multi-step rollout training, together with a jointly trained action-readout variant and ActSWM with its fixed readout. All variants use the same offline step-drift protocol and are evaluated under both recorded and all-zero future actions; no environment interaction is involved.

\noindent\textbf{Metrics.} For each sampled window, the first $H$ steps are encoded as context. Let $K_{\mathrm{eval}}$ denote the evaluation rollout horizon. From the same context, we perform two $K_{\mathrm{eval}}$-step latent rollouts: one conditioned on the recorded future action sequence, producing $\hat z^{\mathrm{gt}}_{t+1:t+K_{\mathrm{eval}}}$, and another conditioned on an all-zero action sequence, producing $\hat z^{0}_{t+1:t+K_{\mathrm{eval}}}$. For each $k\in\{1,\ldots,K_{\mathrm{eval}}\}$, we compute the three quantities in Eq.~\ref{eq:step-drift-metrics}:
\begin{subequations}
\label{eq:step-drift-metrics}
\begin{alignat}{2}
&s^{\mathrm{gt}}_k
&&= \cos\bigl(\hat z^{\mathrm{gt}}_{t+k}, z_{t+k}\bigr),
\label{eq:step-drift-gt}\\
&s^{0}_k
&&= \cos\bigl(\hat z^{0}_{t+k}, z_{t+k}\bigr),
\label{eq:step-drift-zero}\\
&\Delta_k
&&= s^{\mathrm{gt}}_k-s^{0}_k.
\label{eq:step-drift-gap}
\end{alignat}
\end{subequations}
Here $s^{\mathrm{gt}}_k$ measures similarity under the recorded actions, $s^{0}_k$ measures similarity under all-zero actions, and $\Delta_k$ is the action gap. Context Collapse is indicated by high $s^{\mathrm{gt}}_k$ but a small $\Delta_k$: the model predicts plausible futures while failing to make actions change those futures. The complete step-drift protocol restates these metrics in Appendix~\ref{app:step-drift-protocol}.

\noindent\textbf{Results.} Longer context and multi-step training improve prediction under recorded actions but leave the rollouts conditioned on recorded and zero actions nearly indistinguishable. A jointly trained readout increases separation at the cost of prediction quality, whereas ActSWM retains high fidelity, reduces similarity under zero actions by over 80\%, and achieves the largest action gap.

\subsection{Q2: Closed-Loop Model-Based Planning}

\noindent\textbf{Task setup.} We test whether the action-sensitive dynamics from Q1 improve closed-loop planning under the same training data and planner. MineStudio evaluation covers torch placement, stone mining, and pillar building, spanning short interactions, sustained control, and sequential actions \citep{MineStudio}. Related open-world systems evaluate policy learning in MineDojo or Craftax \citep{LSImagine,ImprovingTWM}; our reference-tracking protocol instead uses LeWM as the matched backbone baseline. Planning targets are encoded from task-specific reference trajectories described in Appendix~\ref{app:cem-planning}.

\noindent\textbf{Planning objective.} The encoded reference trajectory provides a sequence of visual goals for model predictive control (MPC). Let $K_{\mathrm{plan}}$ denote the planning horizon. At each environment step, given the current context and a reference future latent sequence $z^{\mathrm{goal}}_{t+1:t+K_{\mathrm{plan}}}$, the planner searches over an action-block sequence $\mathbf{a}_{t:t+K_{\mathrm{plan}}-1}$ of length $K_{\mathrm{plan}}$. For any candidate sequence, the world model rolls out $K_{\mathrm{plan}}$ steps and predicts $\hat z_{t+1:t+K_{\mathrm{plan}}}(\mathbf{a})$. We define the planning cost as the average latent distance to the reference future:
\begin{equation}
J(\mathbf{a})
= \frac{1}{K_{\mathrm{plan}}}\sum_{k=1}^{K_{\mathrm{plan}}}
\left\|
\hat z_{t+k}(\mathbf{a}) - z^{\mathrm{goal}}_{t+k}
\right\|_2.
\end{equation}
The optimized action sequence is
\begin{equation}
\mathbf{a}^{*}
= \operatorname*{arg\,min}_{\mathbf{a}_{t:t+K_{\mathrm{plan}}-1}} J(\mathbf{a}).
\end{equation}
We use CEM to minimize $J$, execute the first optimized action block, and replan after the next observation \citep{CEM}. All models share the action space, targets, optimizer, and execution rule; we report success over 20 trials. Full planner settings and success criteria are given in Appendix~\ref{app:cem-planning}.

\begin{figure}[!t]
\centering
\includegraphics[width=0.8\columnwidth]{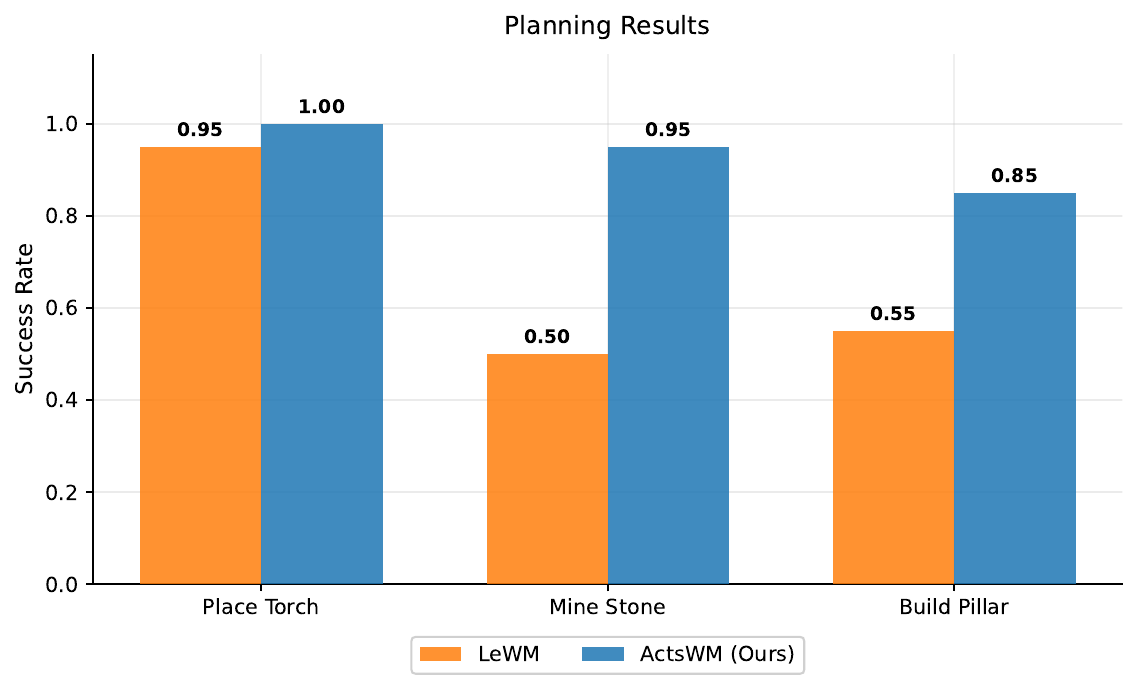}
\caption{Success-rate comparison on Minecraft planning tasks. ActSWM outperforms LeWM on placing a torch, mining a stone block, and building a pillar.}
\label{fig:planning}
\end{figure}

\noindent\textbf{Results.} Relative to the baseline, ActSWM improves success by 5.3\% on torch placement, 90.0\% on stone mining, and 54.5\% on pillar building. The larger gains on sustained and sequential tasks indicate that action-sensitive rollouts improve planning when success depends on distinguishing long action sequences.

\subsection{Q3: Discriminative Action Representations}
\label{sec:local-action-recovery}

\noindent\textbf{Evaluation setup.} We test whether latent dynamics can recover the actions explaining transitions in offline videos from Counter-Strike~2 (CS), Grand Theft Auto~V (GTA), and Apex Legends (Apex). These games cover distinct visual layouts, motion patterns, and interaction rhythms, providing a diverse evaluation of action recovery. Given context and a target frame, CEM searches for an action sequence whose rollout approaches the target, providing an open-loop measure of recoverable action information. We compare direct inverse-dynamics prediction with CEM recovery using LeWM and ActSWM. Latent-action methods such as CoLA-World and LAWM infer task-specific action spaces and therefore are not directly scored against the fixed 14-control vocabulary used here \citep{CoLAWorld,LatentActionWM}. Data processing, context lengths, and the stepwise CEM procedure are detailed in Appendices~\ref{app:cem-eval} and~\ref{app:gameplay-cleaning}.

\begin{algorithm}[t]
\caption{Stepwise CEM for Local Action Recovery}
\label{alg:cem}
\begin{algorithmic}[1]
\REQUIRE World model $M$, context latents $\mathbf{z}_{t-H+1:t}$, target latent $z_{t+\delta}$, action pool $\mathcal{A}$, GT statistics $\mu_{\mathrm{gt}},\sigma_{\mathrm{gt}}$, horizon $K$, CEM params $(N_c,N_e,N_i)$
\ENSURE Recovered action sequence $\hat{\mathbf{a}}_{1:K}$
\STATE $\mathbf{b}\gets[a_{t-H+1},\dots,a_t]$ \COMMENT{action buffer from GT context}
\STATE $\mathbf{e}\gets\mathbf{z}_{t-H+1:t}$ \COMMENT{latent state}
\FOR{$k=1$ \TO $K$}
    \IF{$k=1$}
        \STATE $\mu\gets\mu_{\mathrm{gt}},\;\sigma\gets\sigma_{\mathrm{gt}}$
    \ELSE
        \STATE $\mu\gets\hat a_{k-1},\;\sigma\gets0.5\sigma_{\mathrm{gt}}$
    \ENDIF
    \FOR{$i=1$ \TO $N_i$}
        \IF{$i=1$}
            \STATE Sample $\lfloor N_c/4\rfloor$ candidates from $\mathcal{A}$ and the rest from $\mathcal{N}(\mu,\sigma)$
        \ELSE
            \STATE Sample $N_c$ candidates from $\mathcal{N}(\mu,\sigma)$
        \ENDIF
        \FOR{each candidate $c$}
            \STATE Build 56-dim input: $\mathbf{b}[-3{:}]\oplus c$
            \STATE $\hat z\gets M.\mathrm{predict}(\mathbf{e},\mathrm{action})$
            \STATE $\mathrm{cost}(c)\gets\|\hat z-z_{t+\delta}\|_2^2$
        \ENDFOR
        \STATE $\mathcal{E}\gets\text{top-}N_e\text{ candidates by lowest cost}$
        \STATE $\mu\gets\mathrm{mean}(\mathcal{E}),\;\sigma\gets\mathrm{std}(\mathcal{E})$
    \ENDFOR
    \STATE $\hat a_k\gets\mu$
    \STATE $\mathbf{e}\gets M.\mathrm{predict}(\mathbf{e},\hat a_k)$; append $\hat a_k$ to $\mathbf{b}$
\ENDFOR
\RETURN $\hat{\mathbf{a}}_{1:K}$
\end{algorithmic}
\end{algorithm}

After constructing the full ${K_{\mathrm{rec}}=32}$-step sequence, we compute the local recovery cost by rolling out the world model and comparing the final predicted latent with the target-frame latent:
\begin{equation}
J_{\mathrm{local}}(\hat{\mathbf{a}})
=
\left\|\hat z_{t+K_{\mathrm{rec}}}(\hat{\mathbf{a}})-z_{t+\delta}\right\|_2^2.
\end{equation}

\noindent\textbf{Metrics.} We report three metrics.

\textit{Action Gap.} We compare the CEM recovery cost with random sequences sampled from observed actions:

\begin{equation}
\mathrm{Gap}=J_{\mathrm{rand}}-J_{\mathrm{cem}},
\end{equation}
where $J_{\mathrm{rand}}$ is the random-sequence rollout cost. A large positive value indicates that the model can steer its latent rollout toward the target.

\textit{Key Accuracy.} We compute per-frame binary key accuracy over all 14 key dimensions. Let $b(x)=\mathbf{1}(x>0.5)$ denote thresholding. Then
\begin{subequations}
\label{eq:key-accuracy}
\begin{alignat}{2}
&c_{i,k,d}
&&= \mathbf{1}\!\left[b(\hat a_{i,k,d})=b(a_{i,k,d})\right],
\label{eq:key-correct}\\
&\mathrm{Acc}_{\mathrm{key}}
&&= \frac{1}{N K_{\mathrm{rec}} D_a}\sum_{i,k,d} c_{i,k,d},
\label{eq:key-accuracy-average}
\end{alignat}
\end{subequations}
where $c_{i,k,d}$ indicates a correct control prediction, with sums over $i=1,\ldots,N$, $k=0,\ldots,K_{\mathrm{rec}}-1$, and $d=1,\ldots,D_a$. Here $N=15$ is the number of evaluation windows per game and $D_a=14$ is the number of binary control dimensions.

\textit{Active Accuracy.} Key accuracy alone can be inflated by correctly predicting the dominant zero class for rarely pressed keys. We therefore also report active accuracy, which evaluates recall only on frames where the ground-truth key is active:
\begin{equation}
\mathrm{Acc}_{\mathrm{active}}
= \frac{\sum_{i,k,d} \mathbf{1}\!\left[b(\hat a_{i,k,d})=1 \land b(a_{i,k,d})=1\right]}
{\sum_{i,k,d} \mathbf{1}\!\left[b(a_{i,k,d})=1\right]}.
\end{equation}
This metric directly measures whether the model recovers actions when the human player is actually pressing keys.

Since G-IDM directly outputs action predictions without world-model rollouts, we report only its key accuracy and active accuracy; the Gap metric does not apply. For world-model methods, all three metrics are computed from the CEM-recovered action sequence. Appendix~\ref{app:cem-eval} describes the shared data alignment and evaluation protocol.

\begin{table}[t]
\centering
\caption{Local action-recovery results over 14 binary controls. Gap measures the improvement of CEM over random search. G-IDM predicts actions directly and therefore has no Gap.}
\label{tab:local-action-recovery}
\setlength{\tabcolsep}{4pt}
\resizebox{\columnwidth}{!}{%
\begin{tabular}{llrrr}
\hline
Game & Method & Gap $\uparrow$ & Key Acc.\ $\uparrow$ & Active Acc.\ $\uparrow$ \\
\hline
CS & D2E G-IDM & -- & \textbf{0.913} & 0.000 \\
CS & LeWM & 54.86 & 0.703 & 0.363 \\
CS & \textbf{ActSWM (Ours)} & \textbf{60.02} & 0.743 & \textbf{0.389} \\
\hline
GTA & D2E G-IDM & -- & \textbf{0.709} & 0.000 \\
GTA & LeWM & 19.57 & 0.684 & 0.050 \\
GTA & \textbf{ActSWM (Ours)} & \textbf{252.38} & 0.662 & \textbf{0.761} \\
\hline
Apex & D2E G-IDM & -- & \textbf{0.876} & 0.000 \\
Apex & LeWM & 10.44 & 0.639 & \textbf{0.576} \\
Apex & \textbf{ActSWM (Ours)} & \textbf{172.01} & 0.721 & 0.269 \\
\hline
\end{tabular}%
}
\end{table}

\noindent\textbf{Results.} ActSWM improves the action gap in all three domains, by up to 16.5$\times$, and raises active accuracy by up to 0.711 while maintaining competitive overall key accuracy. G-IDM attains zero active accuracy despite moderate overall accuracy because it mainly predicts the dominant inactive class. The consistent action-gap gains demonstrate stronger action-conditioned steerability.

Across the three evaluations, metrics that explicitly compare action alternatives are more informative about planning utility than future-state similarity or overall key accuracy alone. Q1 establishes this distinction diagnostically, Q2 verifies its consequence in closed-loop control, and Q3 shows that the same action-sensitive representations support offline action inference. The shared mechanism is a latent scoring landscape that changes meaningfully with candidate actions, enabling CEM to rank alternatives rather than merely follow visual context. Thus, rollout separation serves both as a diagnostic of controllability and as a training principle for practical planning. These results establish action sensitivity as more than an auxiliary representation property: it links counterfactual rollout comparison, reliable closed-loop planning, and scalable action labeling from gameplay video.

\section{Conclusion}

We introduced ActSWM to address Context Collapse, where plausible latent rollouts lose sensitivity to action inputs. ActSWM combines multi-step prediction, alternative-action rollout separation, and a fixed action readout to preserve distinguishable, action-recoverable transitions. Across step-drift diagnostics, closed-loop planning, and offline action recovery, ActSWM improves action sensitivity and task success while retaining prediction quality. These results establish action sensitivity as a core requirement for planning-oriented latent world models beyond future-state accuracy.

\bibliography{aaai2027}

\ifshowappendix
\appendix
\setcounter{secnumdepth}{1}

\section{Model Architecture and Training Configuration}
\label{app:training-config}

This section lists the architecture and optimization settings used for the experiments in the main text. Actions are aggregated with \texttt{frameskip=4}, so one world-model step corresponds to four original environment steps. Table~\ref{tab:appendix-model-config} summarizes the shared model and training configuration.

\begin{table}[!ht]
\centering
\caption{Model and training configuration.}
\label{tab:appendix-model-config}
\small
\setlength{\tabcolsep}{4pt}
\renewcommand{\arraystretch}{1.05}
\begin{tabular}{@{}p{0.15\columnwidth}p{0.44\columnwidth}p{0.24\columnwidth}@{}}
\toprule
Category & Item & Value \\
\midrule
Data & frameskip & 4 \\
Data & segment length $H+K$ & 44 \\
Data & split ratio & 90/5/5\% \\
Image & input resolution & 128 px \\
Image & patch size & 16 \\
Encoder & backbone & ViT-tiny \\
Latent & embed dim & 192 \\
Predictor & depth / heads & 6 / 16 \\
Predictor & MLP hidden dim & 2048 \\
Predictor & action conditioning & AdaLN \\
Training & batch size & 32 \\
Training & optimizer & AdamW \\
Training & learning rate & $1\times 10^{-4}$ \\
Training & weight decay & $1\times 10^{-3}$ \\
Training & max steps & 100k \\
Training & precision & mixed 16-bit \\
Training & gradient clipping & 1.0 \\
Training & random seed & 3072 \\
\bottomrule
\end{tabular}
\end{table}

Table~\ref{tab:appendix-loss-config} lists the hyperparameters corresponding to the objective defined in the method section.

\begin{table}[!ht]
\centering
\caption{Loss and objective hyperparameters.}
\label{tab:appendix-loss-config}
\small
\setlength{\tabcolsep}{4pt}
\renewcommand{\arraystretch}{1.05}
\begin{tabular}{@{}p{0.28\columnwidth}p{0.44\columnwidth}p{0.14\columnwidth}@{}}
\toprule
Loss Term & Parameter & Value \\
\midrule
Rollout loss & context length $H$ & 32 \\
Rollout loss & rollout horizon $K$ & 12 \\
SigReg & weight $\lambda_{\mathrm{sig}}$ & 0.09 \\
SigReg & knots / num proj & 17 / 1024 \\
Hinge loss & weight $\lambda_{\mathrm{hinge}}$ & 0.5 \\
Hinge loss & margin $m$ & 0.3 \\
Action readout & weight $\lambda_{\mathrm{readout}}$ & 1.0 \\
Action readout & predicted-transition weight $\alpha_{\mathrm{pred}}$ & 1.0 \\
Action readout & hidden dim & 512 \\
Action readout & trainable & no \\
\bottomrule
\end{tabular}
\end{table}

The implementation retains the configuration key \texttt{idm.stop\_grad=true}. Despite its name, this option only excludes the action readout parameters $\phi_0$ from the optimizer; it does not detach the latent inputs. The readout loss therefore backpropagates through encoded and rollout-predicted transitions to update the encoder and predictor. In the jointly trained readout ablation, $\phi$ is added to the optimizer while all other settings remain unchanged.

\section{Step-Drift Evaluation Protocol}
\label{app:step-drift-protocol}

Table~\ref{tab:appendix-step-drift-config} provides the implementation settings for the step-drift evaluation. Both rollouts use the same context and world model and differ only in their future action inputs.

For each valid window, we encode the first $H$ observations once and autoregressively generate two $K_{\mathrm{eval}}$-step rollouts from that shared context. The recorded-action rollout uses the aligned dataset controls, while the counterfactual rollout replaces every future control with the all-zero vector. At each step, we compute cosine similarity to the same encoded future target and average $s_k^{\mathrm{gt}}$, $s_k^0$, and $\Delta_k=s_k^{\mathrm{gt}}-s_k^0$ over all valid windows. Windows crossing an episode boundary or invalid-frame marker are excluded. All model variants use this identical sampling and aggregation procedure.

\begin{table}[!ht]
\centering
\caption{Step-drift evaluation configuration.}
\label{tab:appendix-step-drift-config}
\small
\renewcommand{\arraystretch}{1.05}
\begin{tabular}{@{}lr@{}}
\toprule
Item & Value \\
\midrule
context length $H$ & 32 \\
evaluation horizon $K_{\mathrm{eval}}$ & 32 \\
frameskip & 4 \\
batch size & 64 \\
max sampled windows & 5000 \\
effective count per step & 1280 \\
random seed & 1234 \\
\bottomrule
\end{tabular}
\end{table}

Table~\ref{tab:appendix-step31-summary} reports the step-31 prediction quality and action-gap values.

The short-context and high-resolution rows isolate context length and input resolution, while the \texttt{+Rollout} rows use multi-step training. The jointly trained readout ablation optimizes the readout together with the world model; ActSWM instead uses the frozen readout described in Appendix~\ref{app:training-config}. Thus, the reported differences arise from the specified model or objective change rather than from the evaluation protocol.

\begin{table}[!ht]
\centering
\caption{Step-31 prediction quality and action gap. GT denotes the rollout conditioned on dataset actions, and Zero denotes the rollout conditioned on all-zero actions.}
\label{tab:appendix-step31-summary}
\small
\setlength{\tabcolsep}{3pt}
\renewcommand{\arraystretch}{1.05}
\begin{tabular}{@{}p{0.55\columnwidth}rrr@{}}
\toprule
Method & GT & Zero & Gap \\
\midrule
LeWM, $H=3$, 128px & 0.239 & 0.244 & -0.005 \\
LeWM, $H=32$, 128px & 0.816 & 0.815 & 0.001 \\
LeWM, $H=3$, 256px, +Rollout & 0.810 & 0.810 & 0.000 \\
LeWM, $H=32$, 128px, +Rollout & 0.972 & 0.970 & 0.002 \\
LeWM, +Rollout, +Joint Readout & 0.698 & 0.106 & 0.592 \\
ActSWM, ours & 0.923 & 0.163 & 0.760 \\
\bottomrule
\end{tabular}
\end{table}

\section{CEM Planning and Task Setup}
\label{app:cem-planning}

This section provides the implementation details for the closed-loop planning experiment in Q2. LeWM and ActSWM use the same planner, action-block library, sampling distributions, target latent sequences, and execution rule; only the world model used for rollout scoring differs. Table~\ref{tab:appendix-cem-config} lists the shared planning configuration.

\begin{figure*}[t]
\centering
\includegraphics[width=0.95\textwidth]{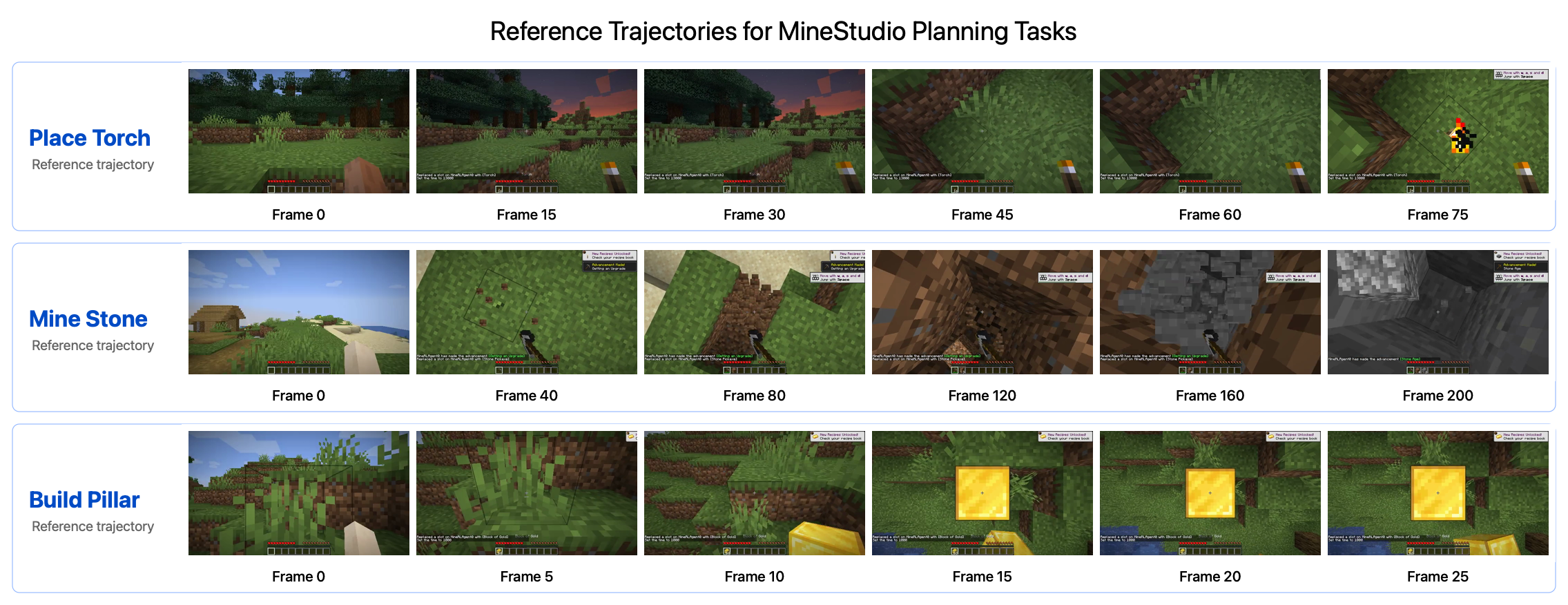}
\caption{Reference trajectories for the three Minecraft planning tasks. Each row shows the target video segment for one task and marks the selected reference frames; these clips are encoded into target latent-embedding sequences for planning.}
\label{fig:reference-tasks}
\end{figure*}

\begin{table}[!ht]
\centering
\caption{Global CEM/MPC planning configuration.}
\label{tab:appendix-cem-config}
\small
\setlength{\tabcolsep}{4pt}
\renewcommand{\arraystretch}{1.05}
\begin{tabular}{@{}p{0.48\columnwidth}p{0.40\columnwidth}@{}}
\toprule
Item & Value \\
\midrule
CEM candidates & 512 \\
CEM iterations & 6 \\
elite fraction & 0.125 \\
rollout batch size & 512 \\
frameskip & 4 \\
action-block library & Table~\ref{tab:minecraft-action-library} \\
planning horizon $K_{\mathrm{plan}}$ & 12 \\
execution & MPC, execute 1 chunk \\
camera dims & Gaussian sampling \\
button dims & Bernoulli sampling \\
\bottomrule
\end{tabular}
\end{table}

\paragraph{Candidate action library.}
In the Minecraft planning experiments, CEM searches over task-specific composite action chunks rather than unconstrained raw actions. Each action chunk contains $4$ raw environment steps. Unless otherwise specified, the same atomic action is repeated for all $4$ steps. Pitch and yaw denote camera movements, while attack, use, jump, forward, and back denote pressed buttons. The prior weights are used to sample candidate chunks before CEM updates the search distribution using elite samples.

\begin{table}[!ht]
\centering
\caption{Task-specific candidate action chunks used by the CEM planner in Minecraft planning experiments.}
\label{tab:minecraft-action-library}
\small
\setlength{\tabcolsep}{3pt}
\renewcommand{\arraystretch}{1.05}
\begin{tabular}{@{}p{0.20\columnwidth}p{0.23\columnwidth}p{0.38\columnwidth}r@{}}
\toprule
Task & Chunk & Action composition & Weight \\
\midrule
Mine Stone & look down & pitch $+8^\circ$ & 0.15 \\
& mine down & attack + pitch $+82^\circ$ & 0.55 \\
& forward & forward & 0.20 \\
& back & back & 0.10 \\
\addlinespace[2pt]
Place Torch & turn right & yaw $+3^\circ$ & 0.20 \\
& forward & forward & 0.20 \\
& look down & pitch $+6^\circ$ & 0.20 \\
& wait & no-op & 0.15 \\
& place & use + pitch $+28^\circ$ & 0.25 \\
\addlinespace[2pt]
Build Pillar & stack action & look down, look down, jump, use + pitch $+45^\circ$ & 0.50 \\
& forward & forward & 0.15 \\
& back & back & 0.15 \\
& turn left & yaw $-5^\circ$ & 0.10 \\
& turn right & yaw $+5^\circ$ & 0.10 \\
\bottomrule
\end{tabular}
\end{table}

Table~\ref{tab:appendix-task-config} reports the task-specific success condition, observation mask, planning budget, environment-step budget, and success count. All tasks use $K_{\mathrm{plan}}=12$.

The \emph{Trigger} column specifies the success signal: torch placement and stone mining use environment events, whereas pillar building is judged from the completed structure. The two values in \emph{Plan--env} are the planning and environment-step budgets, respectively. Each method is evaluated on the same 20 initial conditions and reference trajectories.

\begin{table}[!ht]
\centering
\caption{Task-level settings and success counts.}
\label{tab:appendix-task-config}
\small
\renewcommand{\arraystretch}{1.08}
\resizebox{\columnwidth}{!}{%
\begin{tabular}{@{}lllccc@{}}
\toprule
Task & Trigger & Mask & Plan--env & LeWM & ActSWM \\
\midrule
Place Torch & \texttt{use\_item:torch} & mining & 50--100 & 19/20 & 20/20 \\
Mine Stone & \texttt{mine\_block} & mining & 150--300 & 10/20 & 19/20 \\
Build Pillar & \texttt{human\_label} & none & 50--40 & 11/20 & 17/20 \\
\bottomrule
\end{tabular}
}
\end{table}

\section{Cross-Game CEM Evaluation Protocol}
\label{app:cem-eval}

This section details the shared evaluation protocol for the three-game action-recovery experiment (Q3). We sample 15 episode-aware windows per game, each spanning $K_{\mathrm{rec}}=32$ model steps with frameskip 4 and a target offset of $\delta=128$ raw frames, equal to 32 model steps. Inputs are resized to $128\times128$, and actions are represented by 14 binary controls using the 56-dimensional four-step sliding-window encoding. The controls comprise 10 buttons and four discretized camera directions. ActSWM uses $H=32$, whereas LeWM uses its native context length $H=3$.

For each recovered step, CEM uses 1024 candidates, 128 elites, and 8 iterations. The first step is initialized from $\mathcal{N}(\mu_{\mathrm{gt}},\sigma_{\mathrm{gt}})$; subsequent steps are centered at the previous recovered action with standard deviation $0.5\sigma_{\mathrm{gt}}$. The first CEM iteration mixes 25\% empirical action-pool samples with 75\% Gaussian samples to anchor the search in valid key combinations; subsequent iterations sample purely from the Gaussian to refine around the elite mean. The random baseline draws an i.i.d.\ sequence from the empirical action pool. All experiments are run with \texttt{--human-guide} (other initialization flags such as \texttt{--gt-dist} and \texttt{--zero-init} are not used), together with \texttt{--img-size 128} to ensure uniform input resolution. We report action gap, key accuracy, and active accuracy; D2E G-IDM is evaluated only on the two accuracy metrics because it does not use world-model search.

Evaluation windows are sampled with episode-boundary awareness: a candidate start index is valid only if neither the start frame nor the target frame crosses an episode boundary, both frames belong to the same episode, and both are marked valid. From the valid starts, 15 windows are randomly sampled with seed 42.

All methods are evaluated on the same cleaned windows and binary key dimensions. For world-model methods, key accuracy is computed from the CEM-recovered action sequence. For D2E G-IDM, it is computed directly from the predicted action labels.

\section{Fixed Action-Readout Separation Argument}
\label{app:fixed-readout-separation}

This section expands the separation argument used in the method section. Let $u_i$ and $u_j$ be two latent transitions with corresponding actions $a_i$ and $a_j$. Assume the fixed action readout $q_{\phi_0}$ is locally $L$-Lipschitz on the region containing these transitions:
\begin{equation}
\|q_{\phi_0}(u_i)-q_{\phi_0}(u_j)\| \le L\|u_i-u_j\|.
\end{equation}
If the readout recovers the two actions with errors bounded by $\epsilon_i$ and $\epsilon_j$,
\begin{equation}
\|q_{\phi_0}(u_i)-a_i\|\le \epsilon_i,
\quad
\|q_{\phi_0}(u_j)-a_j\|\le \epsilon_j,
\end{equation}
then the triangle inequality gives
\begin{equation}
\|a_i-a_j\|
\le
\epsilon_i+
\|q_{\phi_0}(u_i)-q_{\phi_0}(u_j)\|
+\epsilon_j.
\end{equation}
Combining this inequality with Lipschitz continuity yields
\begin{equation}
\|u_i-u_j\|
\ge
\frac{\|a_i-a_j\|-\epsilon_i-\epsilon_j}{L}.
\end{equation}
Therefore, if different actions are decoded accurately by a fixed readout, their latent transitions must remain separated by a positive margin whenever $\|a_i-a_j\|>\epsilon_i+\epsilon_j$. This argument does not claim that random initialization is optimal; rather, it explains why freezing the readout provides a stable constraint on latent transition representations. A jointly trained readout does not provide the same constraint because $q_\phi$ can move its decision boundary as the representation changes, reducing action-prediction loss without necessarily increasing separation between transitions associated with different actions.

\section{Gameplay Training Data Cleaning Pipeline}
\label{app:gameplay-cleaning}

This section describes the alignment and cleaning pipeline used for paired gameplay recordings in the cross-game evaluation. It ingests MCAP action streams and corresponding MKV videos and produces HDF5 clips with ground-truth 14-dimensional binary control labels. Public videos without native action logs follow the separate construction pipeline in Appendix~\ref{app:data-pipeline} before model-based action annotation.

\begin{figure*}[t]
\centering
\includegraphics[width=0.92\textwidth]{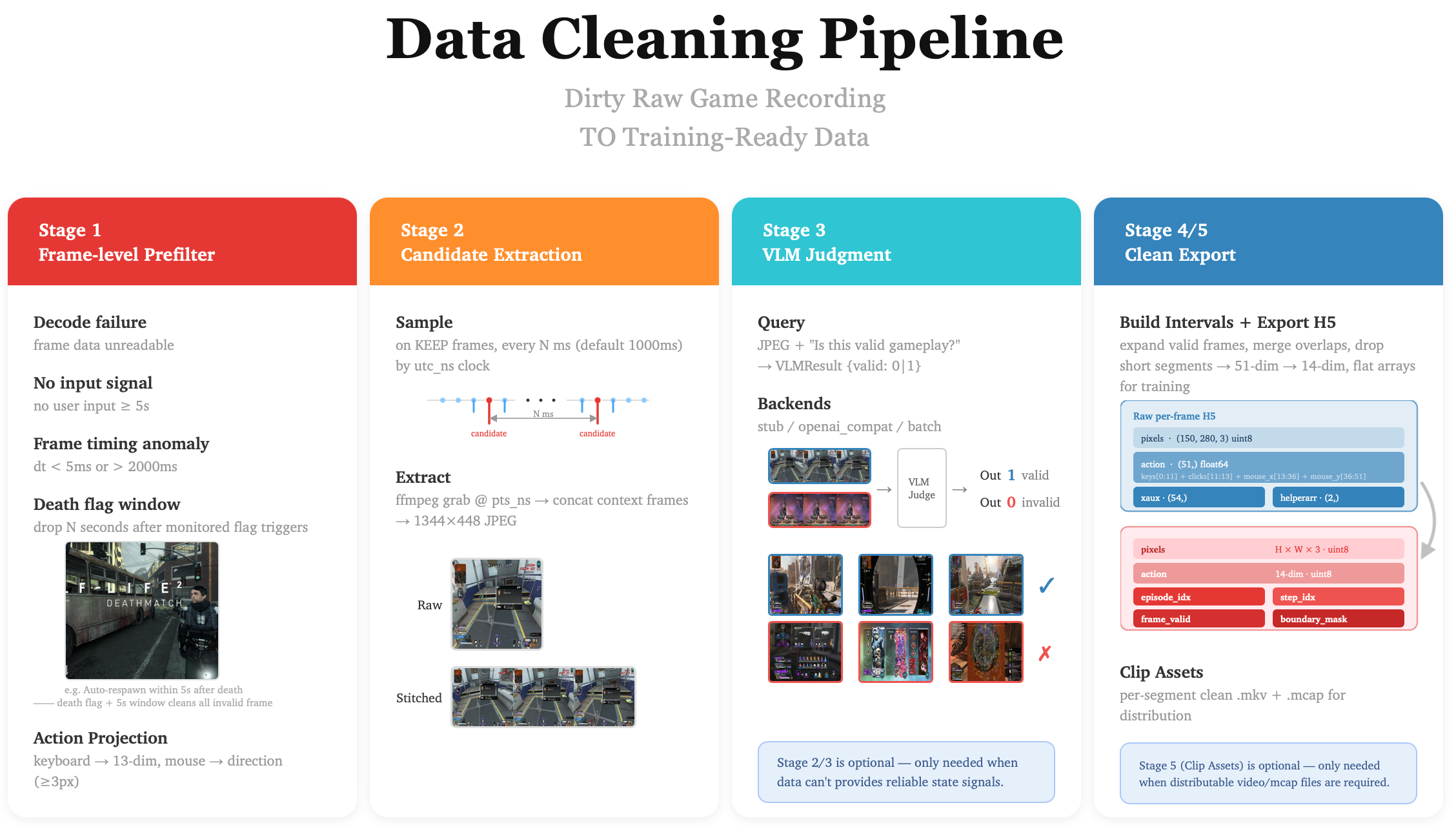}
\caption{Overview of the gameplay data cleaning pipeline.}
\label{fig:cleaning-pipeline}
\end{figure*}

\begin{figure*}[!t]
    \centering
    \includegraphics[width=0.92\textwidth]{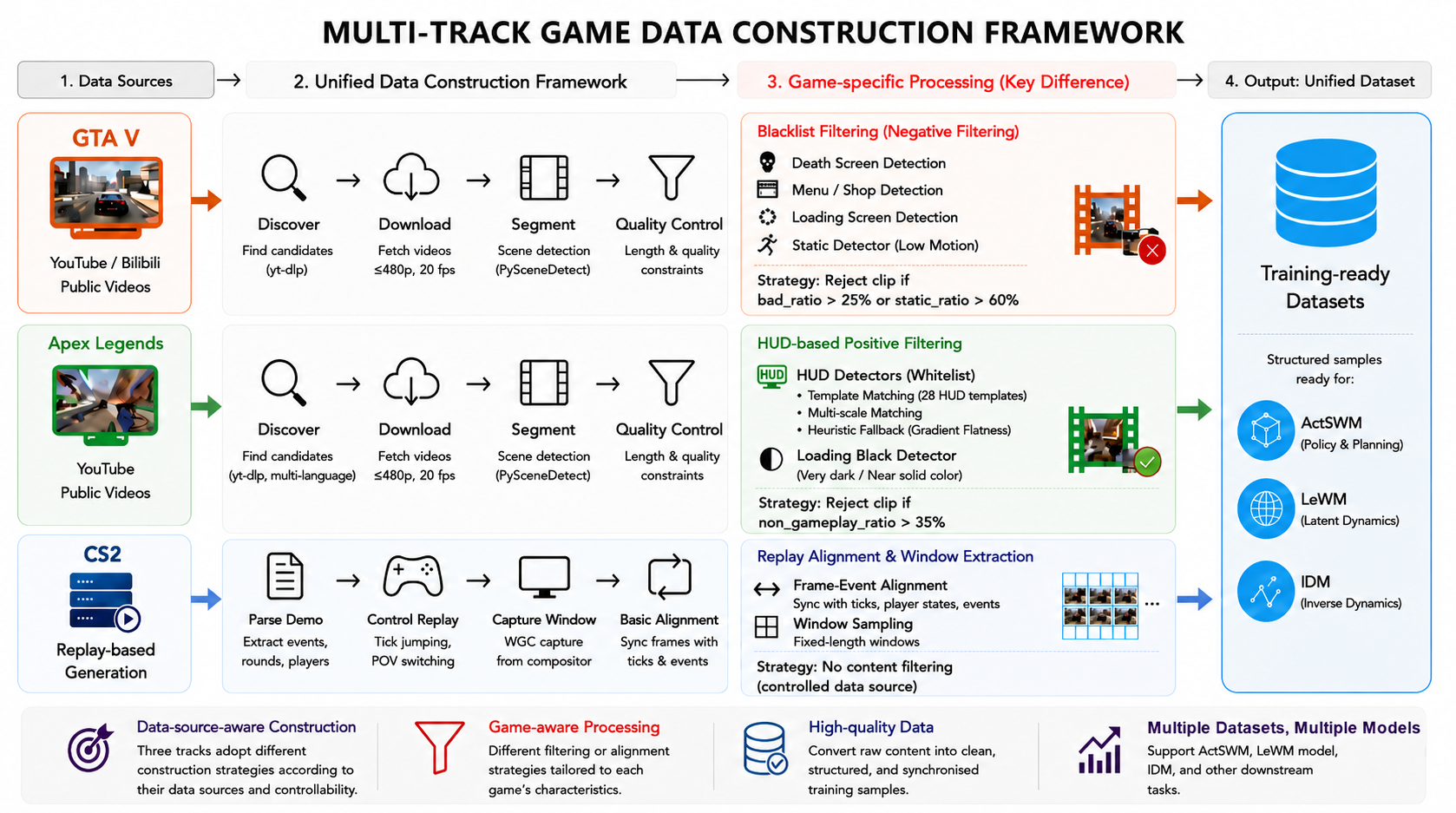}
    \caption{Multi-track game data construction framework. GTA V and Apex share web-video discovery, segmentation, and quality-control stages but use blacklist and HUD-whitelist filtering, respectively. CS2 instead uses controlled demo replay, WGC capture, and frame--event alignment. All pipelines produce a unified training-ready data format.}
    \label{fig:data-pipeline}
\end{figure*}

\paragraph{Stage 1: Rule-based prefiltering.} Raw action streams from the input MCAP are processed frame-by-frame. Frames where keyboard or mouse signals fail to decode, contain no input events, or exhibit timestamp anomalies are discarded. Mouse movements are converted to binary turn/look controls via a threshold of 3 pixels per frame. For sources with explicit death records (e.g., CS deathmatch), frames within a fixed time window after each death event are also removed at this stage. The prefiltered frames are segmented into sub-episodes by splitting at PTS gaps exceeding 150\,ms, discarding segments shorter than 200\,ms. Valid frames are then mapped via the action projection module to a 14-dimensional binary control vector (buttons: W, A, S, D, Shift, Ctrl, Fire, Aim, Reload, Jump; camera: Turn\_L, Turn\_R, Look\_U, Look\_D). Frames within 3 frames of a segment boundary are marked as \texttt{frame\_valid=0} to prevent evaluation windows from crossing PTS-discontinuity boundaries.

\paragraph{Stage 2--3: Candidate extraction and VLM judging.} For domains where non-gameplay states (menus, loading screens, spectator views) are common, candidate frames are extracted from the prefiltered stream at a sampling interval of 1\,s and submitted to a VLM judge. The VLM classifies each frame as valid gameplay or non-gameplay using a domain-specific prompt, operating in either online mode (per-frame API calls with thread-pool concurrency) or batch mode (batch file submission for reduced cost). Consecutive valid frames separated by gaps under 150\,ms are merged into contiguous intervals; intervals shorter than 200\,ms are discarded. The resulting valid intervals define the final export boundaries.

\paragraph{Stage 4--5: HDF5 export and post-processing.} Frames within the validated intervals are read from the source video via ffmpeg (decoded at the source frame rate with pixel format conversion) and written into an HDF5 file containing six datasets: \texttt{pixels} (video frames), \texttt{action} (14-dim binary key labels), \texttt{episode\_idx}, \texttt{step\_idx}, \texttt{frame\_valid}, and \texttt{boundary\_mask}. Episode boundaries are marked to prevent cross-episode window sampling during evaluation. An optional Stage~5 crops and exports cleaned video and MCAP segments for downstream inspection; for H5-mode pipelines, Stage~5 outputs per-stage summary statistics instead.

\section{Model-annotated Dataset Construction Pipeline}
\label{app:data-pipeline}

Figure~\ref{fig:data-pipeline} presents three game-specific data pipelines: GTA~V, Apex Legends, and Counter-Strike~2 (CS2). Although their sources differ, all three pipelines convert raw content into clean, structured first-person gameplay clips suitable for subsequent model annotation. We factor the work into two algorithms. Algorithm~\ref{alg:web-pipeline} covers the shared public-video construction skeleton of GTA and Apex; the \emph{key difference} is the domain filter $F_d$---negative blacklist filtering for GTA vs.\ positive HUD filtering for Apex. Algorithm~\ref{alg:cs2-pipeline} covers CS2, which generates video from structured demos via tick-aligned engine control, GPU-compositor capture, and frame--event alignment (no content filtering; the source is already controlled). Paired recordings used for quantitative evaluation retain native controls through Appendix~\ref{app:gameplay-cleaning}; clips without action logs can instead be annotated with the recovery procedure in Appendix~\ref{app:cem-eval}. In the main text and result tables, CS2 is abbreviated as CS.

\begin{algorithm}[!t]
\caption{Public Video Gameplay Model-annotated dataset Construction Pipeline (GTA~V / Apex Legends)}
\label{alg:web-pipeline}
\begin{algorithmic}[1]
\REQUIRE Domain $d\in\{\mathrm{GTA},\mathrm{Apex}\}$; keywords $\mathrm{kw}$; optional seed URLs; thresholds $\tau_d$
\ENSURE Clean gameplay clips $V_d$ for model annotation (mute, fixed fps/res.)
\STATE $S_d\gets\mathrm{Crawl}(\mathrm{kw})$ \COMMENT{discover source videos}
\STATE $S_d\gets\mathrm{Download}(S_d)$ \COMMENT{optional seeds mainly for GTA}
\STATE $S_d\gets\mathrm{Norm}(S_d)$ \COMMENT{normalize source videos}
\STATE $C_d\gets\mathrm{SceneCut}(S_d)$ \COMMENT{Segment: short clip candidates}
\STATE $\mathcal{B}\gets\{\mathrm{death},\mathrm{menu},\mathrm{load},\mathrm{static}\}$
\FOR{$c\in C_d$}
    \IF{$d{=}\mathrm{GTA}$}
        \STATE $\rho\gets\mathrm{Blacklist}(c;\mathcal{B})$ \COMMENT{negative filtering}
        \STATE $\mathbf{1}_{\mathrm{keep}}\gets[\rho\le\tau_{\mathrm{GTA}}]$ \COMMENT{reject if bad-state ratio high}
    \ENDIF
    \IF{$d{=}\mathrm{Apex}$}
        \STATE $e\gets\mathrm{HUDMatch}_{\mathrm{ms}}(c)$ \COMMENT{positive filtering; multi-scale/layout}
        \STATE $\ell\gets\mathrm{LoadingDetect}(c)$ \COMMENT{reject dark/near-uniform transitions}
        \STATE $\mathbf{1}_{\mathrm{keep}}\gets[e\ge\tau_{\mathrm{Apex}}\land\ell{=}\mathrm{false}]$ \COMMENT{require play HUD; drop loads}
    \ENDIF
    \IF{$\mathbf{1}_{\mathrm{keep}}$}
        \STATE $V_d\gets V_d\cup\{\mathrm{MuteExport}(c)\}$
    \ENDIF
\ENDFOR
\RETURN $V_d$
\end{algorithmic}
\end{algorithm}

\begin{algorithm}[!t]
\caption{Replay-based Gameplay Data Generation Pipeline (CS2)}
\label{alg:cs2-pipeline}
\begin{algorithmic}[1]
\REQUIRE Legal demos $\mathcal{D}$ (\texttt{.dem}/\texttt{.dem.zst}); HLAE/CFG stack; WGC; FFmpeg
\ENSURE Tick-aligned gameplay clips $V_{\mathrm{CS2}}$ for model annotation (mute, fixed fps/res.)
\STATE $M\gets\mathrm{Parse}(\mathcal{D})$ \COMMENT{players, rounds, ticks, kills, account IDs}
\STATE $C\gets\{(p,r,[t_0,t_1])\}$ \COMMENT{player$\times$round alive-only windows}
\FOR{$(p,r,[t_0,t_1])\in C$}
    \STATE $\mathrm{cfg}\gets\mathrm{Jump}(t_0){\circ}\mathrm{Spec}_1(p)$ \COMMENT{seek and switch POV}
    \STATE $\mathrm{cfg}\gets\mathrm{cfg}{\circ}\mathrm{StripHUD}{\circ}\mathrm{Quit}(t_1)$
    \STATE $\mathrm{Launch}(\mathrm{HLAE},\mathrm{Play}(M,\mathrm{cfg}))$ \COMMENT{controlled replay}
    \STATE $f\gets\mathrm{WGC}(\cdot;[t_0,t_1])$ \COMMENT{GPU compositor capture}
    \STATE $c\gets\mathrm{FrameEventAlign}(f,M)$ \COMMENT{align ticks, states, and events}
    \STATE $v\gets\mathrm{MuteExport}(c)$ \COMMENT{\texttt{full\_round.mp4}}
    \STATE $V_{\mathrm{CS2}}\gets V_{\mathrm{CS2}}\cup\{v\}$
\ENDFOR
\RETURN $V_{\mathrm{CS2}}$
\end{algorithmic}
\end{algorithm}

Algorithm~\ref{alg:web-pipeline} formalizes the shared stages of video discovery, download, segmentation, domain-specific filtering, and export. The filtering strategy is the key difference: GTA uses \emph{negative} blacklist filtering over death, menu, loading, and static states, rejecting clips with a high unwanted-state ratio. Apex instead uses \emph{positive} multi-scale HUD whitelist matching because motion alone is unreliable in combat. Algorithm~\ref{alg:cs2-pipeline} generates tick-aligned synchronized training samples from structured demos via engine control, WGC GPU-compositor capture rather than GDI framebuffer capture, and $\mathrm{FrameEventAlign}$ (implementation details in the supplementary software).

The filtering rules are high-throughput construction heuristics rather than standalone detector benchmarks. For GTA~V, clips are rejected when the bad-state ratio exceeds 25\% or the static ratio exceeds 60\%; approximately 40\% of segmented footage is retained, with a target of roughly 250 hours. Apex uses 28 multi-scale HUD templates, rejects clips whose non-gameplay ratio exceeds 35\%, and disables static-motion rejection. CS2 uses controlled replay without content filtering and targets roughly 100 hours of synchronized first-person video. We manually inspected sampled retained and rejected clips to verify the resulting data quality.

\fi
\end{document}